\RequirePackage{amsmath}
\RequirePackage{fix-cm}

\documentclass{article}  
\usepackage{graphicx}

\usepackage[round]{natbib}

\usepackage[utf8]{inputenc}

\usepackage{authblk}

\usepackage{soul}

\usepackage{url}

\usepackage[a4paper, total={6.5in, 8.25in}]{geometry}

\begin{document}

\title{A combined approach to the analysis of speech conversations in a contact center domain}

\author[1]{Andrea Brunello\footnote{Corresponding author: andrea.brunello@uniud.it}}
\author[2]{Enrico Marzano}
\author[1]{Angelo Montanari}
\author[3]{Guido Sciavicco}

\affil[1]{Dept. of Mathematics, Computer Science, and Physics, University of Udine, Udine, Italy}
\affil[2]{Research and Development Department, Gap S.r.l.u., Udine, Italy}
\affil[3]{Dept. of Mathematics and Computer Science, University of Ferrara, Ferrara, Italy}

\date{}

\maketitle

\begin{abstract}
The ever more accurate search for deep analysis in customer data is a really strong technological trend nowadays, quite appealing to both private and public companies. This is particularly true in the contact center domain, where speech analytics is an extremely powerful methodology for gaining insights from unstructured data, coming from customer and human agent conversations. In this work, 
we describe an experimentation with a speech analytics process for an Italian contact center, that deals with call recordings extracted from inbound or outbound flows. First, we illustrate in detail the development of an in-house speech-to-text solution, based on Kaldi framework, and evaluate its performance (and compare it to Google Cloud Speech API). 
Then, we evaluate and compare different approaches to the semantic tagging of call transcripts, ranging from classic regular expressions to machine learning models based on ngrams and logistic regression, and propose a combination of them, which is shown to provide a consistent benefit. 
Finally, a decision tree inducer, called  J48S, is applied to the problem of tagging. 
Such an algorithm is natively capable of exploiting sequential data, such as texts, for classification purposes. The solution is compared with the other approaches and is shown to provide competitive classification performances, while generating highly interpretable models and reducing the complexity of the data preparation phase. The potential operational impact of the whole process is thoroughly examined.
\end{abstract}

\section{Introduction}
\label{intro}

Nowadays, more and more companies strive to extract relevant knowledge regarding their business. Although such data may in fact be an important source of strategic information, it is sometimes stored in an unstructured and hard to exploit form, as, for instance, in the case of data generated from text or audio flows. Specifically, the ability to analyze conversational data plays a major role in contact centers, where the core part of the business still focuses on the management of oral interactions \citep{saberi2017past}.

The analysis of texts originating from oral conversations has to face two main challenges: $(i)$ the intrinsic characteristics of conversational dynamics, which are different from those of written texts, and $(ii)$ the presence of a transcription process, which introduces a bias between the spoken interaction and its written form. 
The combined effect of these two factors asks for a specific treatment, different from the one exploited for written compositions. 

\smallskip

In this paper, we focus on the specific case of a real company, to root our work on actual issues emerging from practice, but the solutions we propose are of general validity. 

We analyze a dataset of recorded anonymized agent-side call conversations produced by a wide range survey campaign made on a large part of the Italian population by Gap S.r.l.u., an Italian business process outsourcer specialized in contact center services. 
Such an analysis is extremely valuable to the company, as it may be easily integrated in a broader,  existing decision support system it developed during the last years. 
It should be noted that, to the best of our knowledge, this work is one of the few to tackle with speech analytics applied to a real business case in Italian language. 

\smallskip

The ultimate goal of the work is the development of a reliable speech analytics solution with a real operational impact on company processes, such as, e.g., in assessing the overall training level of human employees. In particular, we aim at determining whether it is possible to develop a full-fledged speech analytics system at a relative low industrial cost by relying on interpretable models only, that can be validated by a domain expert before being put into production. As we shall see, this led us to neglect two approaches that are commonly used in natural language processing, i.e., neural networks and Conditional Random Fields (CRFs).

As a preliminary step, we check the feasibility of building a proprietary transcription model trained on a relatively small dataset. In that respect, the comparison with a commercial cloud transcription solution is  mandatory to evaluate its costs and performances.

The second main challenge concerns the tagging of conversation transcripts. We investigate potentialities and limitations of a traditional, ngram-based machine learning solution. More precisely, we try to establish whether an approach based on machine learning may be reliable enough, and more efficient than a linguistic solution based on domain expert-defined regular expressions. As a by-product, we assess the effectiveness of a combined approach to tagging that mixes the two. 

Finally, on the basis of the results obtained from the application of the linguistic and classical machine learning approaches, we assess the impact on the text analysis process of a solution based on a decision tree induction algorithm capable of seamlessly handling sequential data \citep{j48s}. 
The outcome of the evaluation is that such a solution has at least two major key benefits: the simplification of the data preparation phase and the high interpretability of the final model which, as we have already pointed out, is one of our main goals.

\smallskip

Although the considered case study dates back to 2017, we believe it to be still interesting for the community, as it shows how, based on a cost-free solution that does not even require any advanced or dedicated hardware, it is possible to develop a full-fledged in-house system for the transcription and the analysis of voice recordings. 

\smallskip

The rest of the paper is organized as follows.
Section \ref{ccdomain} introduces the contact center domain.
Section \ref{sec:relatedw} provides an account of related work on the analysis of call interactions in contact centers.
Section \ref{sec:related} outlines the general framework of our speech analytics process, describing the specific task which we focus on, that is, the semantic tagging of conversations recorded in the context of a specific outbound survey. Section \ref{sec:transcribing} deals with the first part of the analytics workflow, namely, the transcription of phone conversations. 
Section \ref{sec:tagging} focuses on the second phase of the analysis, which consists of the semantic tagging of telephone conversation transcripts. 
Section \ref{sec:decisiontree} focuses on J48S, a decision tree model based on C4.5 \citep{Quinlan:1987:SDT:50007.50008}, which handles sequential data, such as textual data, for classification purposes. Section \ref{busimpact} discusses the operational impact of the proposed speech analytics solution. The last section provides an assessment of the work done and outlines future work directions.

\section{The contact center domain}
\label{ccdomain}
In this paper, we focus on contact centers for front office business process outsourcing, that is, the contracting of a specific business task, like a specific front office activity, to a third-party service provider. Telephone call centers, being an integral part of many businesses, are an increasingly important element of today's business world. In particular, they act as a primary customer-facing channel for firms in many different industries, and they employ millions of operators across the globe.

At its core, a call center consists of a set of resources, typically personnel, computers, and telecommunication equipment, which enable the delivery of services via the telephone. The common work environment is given by a very large room, with numerous open-space workstations, where employees, called \emph{operators} (or \emph{agents}), equipped with headphones, sit in front of computer terminals, providing teleservices to customers on the phone. A current trend, made possible by IT advancements, is the extension of call centers into multi-channel \emph{contact centers}, where the \lq\lq phone operator\rq\rq{} role of the employees is complemented by services offered via other media, such as, for instance, email, chat, and web pages \citep{telrew}.

As it operates, a large contact center generates vast amounts of data, which can be split in two classes: \emph{operational} and \emph{service data}. The former concerns the technical information of a call (phone number, the agent who has served the call, possible call transfers, timestamps, and so on), by which a detailed history of each call that enters the system can, in principle, be reconstructed. The latter is related to the particular service for which the call has been made, e.g., in case of an outbound survey service, data would include all the answers given by the interviewed person.

\smallskip

Contact centers can be categorized with respect to different dimensions. As an example, they can be classified on the basis of the functions they provide (customer services, telemarketing, emergency response, help-desk, order taking). 
Among all the dimensions according to which we can classify them, the distinction between \emph{inbound} and \emph{outbound} activities is of primary importance.

\emph{Inbound contact centers} handle incoming traffic by answering to calls received from the customers, as in the case of helpdesks. Note that inbound operations may generate outgoing calls, as in the case of callback services, i.e., outbound calls made to high-value customers who have abandoned their (inbound) calls before being served, because of, for instance, long waiting times.

\emph{Outbound contact centers} handle outgoing calls, which are made by the contact center. Such calls may be associated with telemarketing initiatives or surveys. Typically, agents involved in this kind of operations follow a pre-defined \emph{script}, which tells them precisely how to manage each call, e.g., how to announce themselves to the called person, how to carry on the call, how to answer to possible questions, and so on.

In both inbound and outbound operations, the contact center agency establishes with its clients a \emph{Service Level Agreement} (SLA), that is, a quality level which has to be maintained for incoming calls, or a set of goals that must be fulfilled during an outbound campaign.

\smallskip

In the last decade, the amount and the importance of data collected by contact centers has had a sensible increase. This led to the introduction of even more sophisticated analytical tools in the ever evolving contact center IT infrastructures. This is the case, for instance, with decision support systems, which are a reliable tool that supports the operational staff in managing services and agents. Moreover, they may lead to the discovery of important insights, deliverable to the companies' core businesses. This happened to Gap S.r.l.u., which has developed, as a proprietary solution, a decision support system, built on a comprehensive data warehouse, able to perform a variety of descriptive and prescriptive analytics tasks. In such a framework, a speech analytics process represents a strong contribution to the overall analytical power.

\section{An account of related work}\label{sec:relatedw}

As pointed out in the introduction, speech analytics may play an important role in the domain of contact centers. Among the benefits of its use, we would like to mention the possibility to exploit it to distinguish between well-behaved and problematic calls.

Existing solutions can be roughly partitioned into those that operate directly on raw audio data and those that turn them into textual data.

An analysis directly based on raw audio features has been proposed in \citep{atassi2014automatic}, where the authors look for specific sound patterns which can be interpreted as, for instance, turn taking, hesitation, or voice overlap. Similarly, in \citep{pandharipande2012novel}, the authors rely on patterns based on the so-called \emph{speaking rate}, which is expressed as the number of spoken words per minute. Both approaches take into consideration notable beahavioural patterns, emerging from an acoustic analysis, with the goal of obtaining a classification of conversations.

As an alternative, the analysis may rely on textual information. This is the case with \citep{cailliau2013mining, zweig2006automated}. These approaches consist of two main steps.
The first one generates the textual data from the raw audio files by making use of an automatic transcription process. Text mining techniques are then applied to the resulting texts with the final goal of determining the overall quality of the interaction.

A more advanced approach is proposed in \citep{mishne2005automatic,garnier2008callsurf}, which makes use of advanced natural language processing (NLP) techniques, such as \emph{noun group}, \emph{named entity recognition}, and \emph{divergence of corpus statistics}, to analyze the obtained transcriptions in the context of a broader architecture for conversation analysis.

In \citep{DBLP:conf/lrec/BechetMBBEMA12}, the DECODA project is illustrated, which aims at facilitating the development of robust speech data mining tools in the context of contact center evaluation and monitoring. The main contribution is the proposal of a French language corpus originated from the Parisian public transportation call center, which ought to reduce the development cost of speech analytics systems by limiting the need for manual data annotation. Based on such a corpus, several studies have been conducted, including theme identification \citep{DBLP:books/sp/15/EsteveBLMDLMM15,DBLP:conf/interspeech/MorchidLEM13,DBLP:conf/interspeech/MorchidDBLM14,DBLP:conf/slt/ParcolletMLM18}, dialogue classification \citep{DBLP:conf/interspeech/KocoCB12,DBLP:conf/icassp/MorchidDBBLM14}, named entity and semantic concept extraction \citep{DBLP:conf/slt/GhannayCECSLM18}, and speech summarization \citep{DBLP:conf/interspeech/TrioneFB16}.

We conclude by underlining once more that, to the best of our knowledge, no systematic work on speech analytics applied to a real business scenario has been done for the Italian language till now.

\section{The general framework}
\label{sec:related}

In this section, we outline the general structure of the solution that we propose.
The work done is inspired by the
approach followed in \citep{mishne2005automatic,garnier2008callsurf}, and it sets the basis for a flexible and modular framework for conversation analysis in a contact center. 

The system consists of two main modules: the first one is in charge of the automatic transcription of  speech conversations, while the second one analyzes the generated texts. The analysis step actually includes two distinct phases, i.e., a lexical and a logical one.
%, which respectively deal with a domain independent analysis and the .
Such a modular approach allows us to provide specific problem-tied solutions and makes it easier to analyze and compare various methods. As a matter of fact, the proposed solution has enabled us to test and combine different methodologies both for the transcription and the text analysis activities, highlighting the best performing configuration.

We focus our attention on the analysis of agent side calls originated in the context of an outbound survey service, carried out by Gap S.r.l.u. In order to guarantee the privacy of the individual agents, all data have been anonymized. As explained in Section \ref{ccdomain}, agents performing outbound calls typically ought to complete a predefined script in order for a call to be considered successful. Such a rigid structure makes it easier to analyze a conversation and to establish whether the agent has followed all the required steps or not. Although such an analysis may seem trivial, it has very deep practical implications for the company. As a matter of fact, the common practice in the domain of call centers is that the overall quality of phone conversations is manually checked by supervising staff by simply listening to a random number of calls. Such an operation is clearly time-consuming and it requires a lot of  enterprise resources. An automatic analysis module may simplify it a lot by identifying \lq\lq problematic\rq\rq{} calls that require further investigation, thus reducing the time needed for verification and increasing the overall efficacy of the process. Based on the analysis tasks, the supervising staff may then identify the need for further training or specific deficiencies that have to be solved. 

\begin{figure*}[t]
\includegraphics[width=\linewidth]{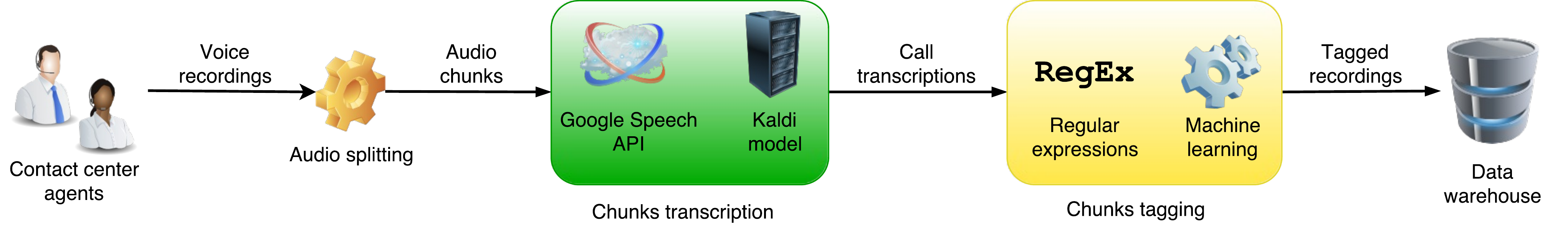}
\caption{The call analysis workflow.}
\label{fig_workflow}
\end{figure*}

The workflow of the process of call analysis is depicted in Figure \ref{fig_workflow}. During each phone conversation, two distinct voice recordings are generated, one for each side of the conversation. The two recordings have exactly the same length, meaning that they are filled with silence phases when the other side is speaking. The first step in the processing of the raw audio is the splitting of the agent-side recording, which is done by considering the pauses (see Section \ref{sec:datakaldi}, proprietary corpus (spontaneous)). The result is a set of \emph{segments}, which are individually passed to the following transcription step. Transcription may be performed either by means of Google Cloud Speech API, or by exploiting an internal, self-trained Kaldi model, as explained in Section \ref{sec:transcribing}. Once the transcripts have been generated, they can be tagged with proper keywords, encapsulating the semantic content of the conversation. As explained in detail in Section \ref{sec:tagging}, transcribed segments may be tagged following two main strategies: the first one relies on {\em regular expressions} defined by domain experts, the other one makes use of machine learning algorithms. Finally, results of the analysis are stored in the company's data warehouse. Data may then be used for reporting purposes, or may form the basis for more advanced tasks such as the firing of automatic rule-based actions.

\section{The transcription of phone conversations}
\label{sec:transcribing}

In this section, we address the problem of speech-to-text transformation of phone calls, with reference to the transcription of agent side phone conversations on a specific outbound survey service. To this end, we analyze and compare the performance of Google Cloud Speech API \citep{googlespeechapi} and Kaldi ASR framework \citep{kaldi}.

\subsection{Transcribing with Google Cloud Speech API}
\label{sec:transcribinggoogle}

In order to obtain a baseline for the transcription performance, against which to evaluate the in-house Kaldi model, we relied on Google Cloud Speech API \citep{googlespeechapi}, a general purpose ASR service that can be used to transcribe conversations in over 80 languages. 

One of the distinctive features of this service is that it can be easily integrated into developer applications. Both online (real time) and offline (batch) transcriptions are supported. Moreover, the service can be tailored to a specific domain, by providing a set of words or phrases that may be pronounced. Such a functionality turns out to be extremely useful, because it allows one to detect personalized names, acronyms, and so on.

We tested the API by developing a Python script that performs a batch transcription of a set of segments (i.e., single phrases), providing also a set of custom words to tailor the transcriptions to the specific outbound service of choice. We considered 339 manually transcribed agent-side segments, taken from successful calls made for the specific outbound service under study (for more information, see Section \ref{sec:transcribing_kaldi}, Paragraph \emph{Proprietary corpus (spontaneous)}, and Table \ref{tab:datasetasr}).

As a result, we obtained a word error rate of 18.70\%. It should be observed that such a result considers common stopwords as well, which  can  mask  the  true  performance  of  automatic speech  recognition  for  the  intended  task.  Thus,  in  an attempt to concentrate on \lq\lq important\rq\rq{} terms, we have also evaluated the transcription system against the same  339  segments,  but  neglecting  Italian stopwords (as listed  in  the  package  NLTK). The result we obtained is not significantly different (in fact, it is slightly better), showing a word error rate of 18.51\%.

\subsection{Transcribing with internal Kaldi model}
\label{sec:transcribing_kaldi}

Kaldi is a free, open-source toolkit for speech recognition, written in C++, and available for both Windows and Unix-like systems for research purposes.
The core library supports the modeling of
arbitrary phonetic-context sizes, and acoustic modeling with Gaussian mixture models (SGMM) as well as deep neural networks. Other than having a large online documentation, the software comes with a variety of \emph{recipes}, i.e., sets of already worked-out scripts to carry out automatic speech recognition (ASR) tasks, that can be tailored to the specific use case.

\subsubsection{Data resources and data preparation}
\label{sec:datakaldi}

For the training of the Kaldi model, we relied on four data sources, each composed of a collection of manually transcribed speech utterances. Two of them, namely, CLIPS and QALL-ME corpora, are freely available for download on their respective websites. The remaining two have been developed inside Gap S.r.l.u., and are proprietary datasets. Table \ref{tab:datasetasr} lists the amount of data contained in each dataset. Note that the overall amount of transcribed instances available for training is rather low (less then 10 hours of speech) and that just about half of them are generated by the company. Nevertheless, as we shall see in Section \ref{sec:transcriptioneval}, they allowed us to reach a good enough level of transcription performances, in a cost-effective manner.

\begin{table}[t]
\centering
\caption{Corpora used for model training and evaluation}
\label{tab:datasetasr}
%\resizebox{0.6\linewidth}{!}{
\begin{tabular}{lllll}
\hline
\textbf{Corpus name}   & \multicolumn{2}{l}{\textbf{\# utterances}} & \multicolumn{2}{l}{\textbf{Recording time}} \\
              & \textbf{training}          & \textbf{test}          & \textbf{training}           & \textbf{test}          \\
\hline
CLIPS         & 1025               & -            & 2h 30m             & -           \\
QALL-ME      & 1208              & -            & 2h 20m             & -            \\
Proprietary (read) & 3467              & -            & 4h 28m             & -            \\
Proprietary (spont) & 201               & 339             & 30m                & 35m           \\
\hline
\end{tabular}
%}
\end{table}

\paragraph{\bf CLIPS.} CLIPS
is a collection of Italian speech and text corpora, freely available for research purposes \citep{clips}. The entire dataset consists of about 100 hours of speech, equally represented by male and female voices, for a total amount of about 300 speakers. Recordings have been captured in 15 Italian cities, selected on the basis of linguistic and socio-economic principles of representativeness. For each city, several kinds of speech have been included, ranging from radio and television broadcasts to telephone conversations. For the purposes of our work, we focused on a subset of the telephone speech corpora, whose audio is stored in monophonic .wav format files, 32 bit, 8kHz, 128Kb/s. Overall, we considered 1025 transcribed utterances, for a total conversation time of about 2 hours and a half. Each utterance is the result of a role play, in which a human person acts as a tourist who is calling a hotel reception to require a service or ask for information. Although using such a small amount of the entire corpus may seem an unusual choice, we purposely did that in order to keep training data as coherent to the domain as possible. The latter consists of phone recordings, that typically exhibit specific characteristics in terms of audio quality, presence of noise, and way to speak. On the contrary, tv, radio, and similar recordings are characterized by a significantly different phraseological articulation and sound quality.

\paragraph{\bf QALL-ME.} The QALL-ME
benchmark consists of 4501 single-speaker utterances \citep{qallme}, declined in several languages. The data were produced by 113 different speakers, mostly native-ones, and equally distributed between males and females. Utterances are divided into spontaneous and read ones. The first are produced by speakers within a scenario which prompts them to ask for information such as, for instance, the opening times of a local museum, the prices of a certain restaurant, and the time a particular movie is showing. The latter are predefined texts read out loudly by speakers. Audio is stored in monophonic .wav format files, 32 bit, 8kHz, 128Kb/s. For the aims of our work, we selected 1208 transcribed spontaneous utterances in Italian language, for a total recording time of about 2 hours and 20 minutes. 

\paragraph{\bf Proprietary corpus (read).} It consists of 3467 transcribed utterances, for a total recording time of about 4 hours and a half. Following the typical workflow of the specific outbound survey service under study, a series of typical, agent-side phrases have been prepared, each corresponding to an utterance. A set of selected, Italian-native contact center agents has then been instructed, through a specific data gathering protocol, to read each of the phrases while wearing a typical phone headset, that has been used to record the voice. Each audio file has then been saved in monophonic .wav format, 32 bit, 8 kHz, 128Kb/s Windows PCM.

\paragraph{\bf Proprietary corpus (spontaneous).} It consists of agent-side recordings taken from successful calls made for the specific outbound service under study. Audio files have been  automatically generated by the company CRM systems, and saved in monophonic .wav format, 32 bit, 8 kHz, 128Kb/s. To integrate such data with that of the other datasets, as a first step, each call recording has been split into segments, on the basis of conversation silences. In order to do that, a Python script relying on the \emph{pydub} library has been coded. Pydub has the ability to split audio considering: $(i)$ \emph{minimum silence length} for the pause to be detected, which, in our case, has been set to 750ms; $(ii)$ \emph{silence threshold}, i.e., the noise cut-off level used in identifying silence, which we empirically set to -34dB; $(iii)$ \emph{keep silence}, by which an amount of silence is kept at the beginning and end of each segment in order to avoid abrupt cuts, set to 450ms. Out of the obtained segments, we kept those lasting at least 3 seconds. The result is a net speaking time of about 65 minutes, over 540 segments.

\subsubsection{Training of the model}
\label{sec:trainingkaldi}
A Kaldi ASR model consists of two parts, a language model and an acoustic model. The first basically defines, by analyzing the training transcripts, the set of phrases that are to be accepted by the recognizer, establishing the likelihood of a word following another. In other words, the language model gives the prior probability of a word sequence. The latter represents the relationship between an audio signal and the linguistic units, such as phonemes, that make up speech.
As we shall discuss, for the training of both models, we relied on the GMM-based \emph{vystadial} Kaldi recipe, which has been adapted for our purposes of building an off-line transcriber for the Italian language. As already pointed out, we decided to neglect more advanced neural network-based acoustic models for a matter of industrial cost: we wanted to determine whether the already present, general purpose system infrastructure could be used to develop the speech analytics solution, without relying on more specialized and expensive hardware.
%Note that, given the scarce amount of training data available, it would not have been feasible to train a deep neural network based acoustic model. 
%, used in the context of a specific outbound survey service.

\paragraph{\bf Language model.} The language model has been built from the above datasets by making use of trigrams. The full vocabulary is made of 1873 distinct Italian words. For common Italian words, the pronunciation dictionary, i.e., the mapping between words and phonemes, has been extracted from the Italian version of the \emph{Festival} text-to-speech software \citep{festival} by means of a custom script. For custom, or domain-oriented words, pronunciations have been manually defined. As a result, a set of 48 phonemes in ARPAbet format has been established, including a list of frequent onomatopoeiae. Although the size of the dictionary may seem small, recall that we are going to analyze agent-side outbound phone conversations, which ought to show a relatively little variance. Thus, also the vocabulary used by the agents is expected to be rather small. As for the handling of out of vocabulary (OOV) words, we chose not to emit them, but rather always return the most likely word among those included in the vocabulary.

\paragraph{\bf Acoustic model.} As mentioned before, given the scarcity of available training data, and the lack of dedicated hardware, for the training of acoustic model we relied on GMMs, neglecting the use of neural networks, which would have required a far greater number of already transcribed utterances, other than increasing the overall development cost. In particular, we relied on a customization of \emph{vystadial\_en} recipe \citep{vystadial}, building a triphone model (LDA + MLLT feature transformation, Maximum Mutual Information objective function), where a triphone is a sequence of three phones and captures the context of single triphone phone.
The model has been trained on a dedicated virtual machine running CentOS 7, and featuring a 3.1GHz quad-core processor as well as 8 GB RAM. Despite the relative modest computing resources, the training process took approximately 5 hours.

\subsubsection{A critical evaluation of the internal model}
\label{sec:transcriptioneval}

We tested the Kaldi approach against the same set of 339 manually transcribed segments, which have been used to evaluate the performance of Google Speech API (see Table \ref{tab:datasetasr}). 
Results indicate a word error rate of 28.77\%, which slightly decreases to 28.33\% when stopwords are neglected.

While it is not surprising that Google Speech surpassed the Kaldi solution in terms of raw transcription accuracy (obtaining 18.70\% and 18.51\% word error rate, respectively), it should be remarked that the in-house model has been trained on a reduced quantity of data, at a relatively low expense for the company. 
Moreover, while Kaldi is a free solution, the use of Google Speech requires an amount of fees per second of conversation to be paid.
Finally, as we shall see in the next section, the quality of the transcriptions generated by Kaldi is good enough to allow for their tagging. 

Even though it is true that the evaluation has been carried out on a single service, for which the Kaldi model has been specifically trained, it can be argued that such a model can be easily extended to deal with other kinds of outbound calls by simply increasing the training set, with a limited effort for the company. Notice that, as the model will become more capable of transcribing heterogeneous conversations, it will be possible to apply it, with increasing success, to the inbound case as well.

\section{Semantic tagging of telephone conversation transcripts}
\label{sec:tagging}

In this section, we concentrate upon the semantic tagging of the conversation transcripts.
%generated by means of the 
%approaches presented in Section 
%\ref{sec:transcribing}. 
Such a process involves the definition of a suitable conceptual schema to represent the keywords within the existing Gap's enterprise-wide data warehouse \citep{mastersthesis}, and the set up of linguistic as well as ngram-based machine learning strategies to guide the tagging process.

\subsection{The database conceptual schema}
\label{sec:dbmodel}

\begin{figure*}[t]
\includegraphics[width=\linewidth]{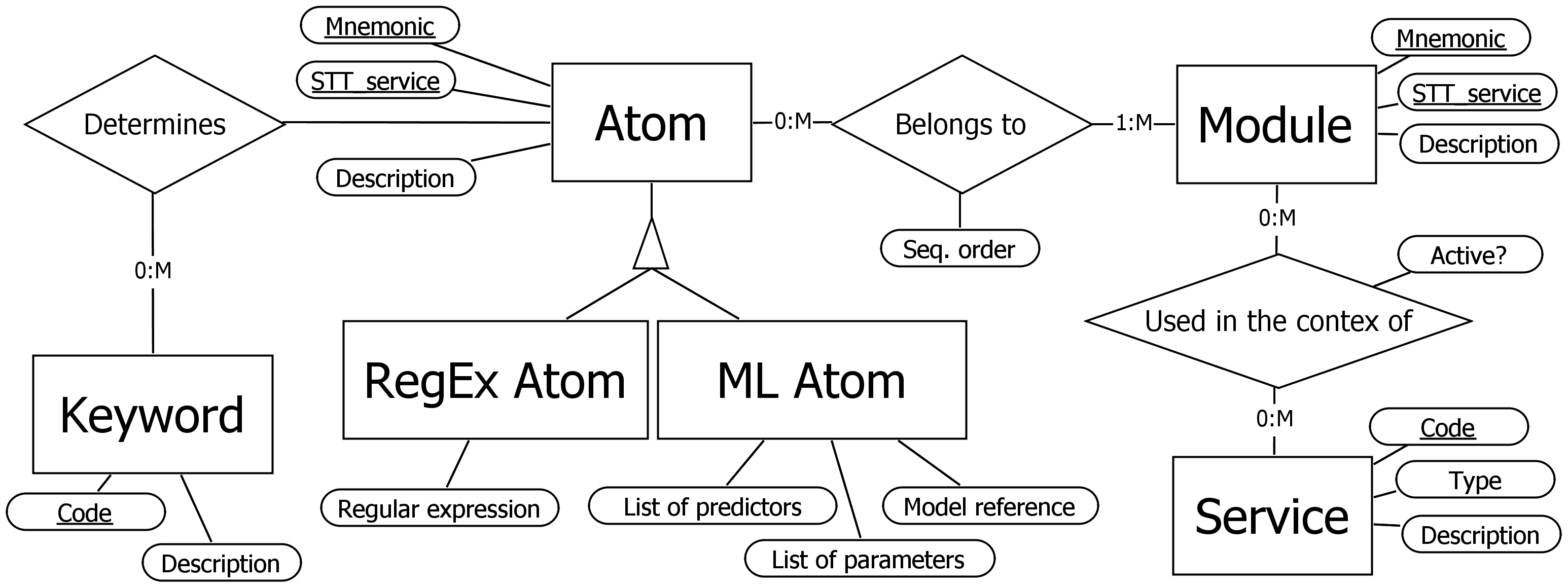}
\caption{A conceptual schema for the tagging process.}
\label{fig_dbmodel}
\end{figure*}

In order to tag the transcribed phone conversations, we first defined a suitable conceptual (database) schema for the encoding of relevant information. The proposed schema is depicted in Figure \ref{fig_dbmodel}, using the Entity - Relationship model.
As a matter of fact, it is an extension to the company's data warehouse schema originally proposed in \citep{mastersthesis}.

Each transcription segment is typically tagged by one or more \emph{keywords}, that is, a word or a combination of words encapsulating a semantic meaning, which expresses what is going on in the corresponding part of the call. Examples of keywords are \emph{first\_survey\_question}, \emph{greetings\_formula}, and \emph{closing\_formula}. A keyword is thus a very general concept that allows one to manage in a uniform way highly heterogeneous phone calls.

A keyword is detected by means of one or more \emph{atoms} (the name suggests the fact that they are the fundamental building blocks of the structure), which can be thought of as strategies to analyze textual segments.  Each atom determines one and only one keyword, and is designed to handle the transcriptions generated by a specific speech-to-text approach, e.g., Kaldi or Google Speech, in the context of a service offered by the company. As an example, each service typically has its own greetings formula, that has to be detected in a dedicated way. 

Atoms are currently specialized into two non overlapping categories: RegEx and Machine Learning (ML), though the generality of the schema makes it easier to add other ones. The former relies on regular expressions designed with the help of domain experts to detect the semantic content of the text. The latter makes use of machine learning models to identify such content. Specifically, an ML-Atom is designed to be as general as possible. As such, it contains: $(i)$ a reference to the specific model that has to be used,  e.g., a logistic regression one, which can be coded directly in an SQL function or into an external script; $(ii)$ a list of parameters to be fed to the model, such as the regression coefficients; and, $(iii)$ a list of the attributes to extract from the textual information to be used as predictors, such as the presence or absence of specific (combinations of) words.

Each atom belongs to a \emph{module}. Modules are basically lists of atoms that have to be checked together, in a specific order. As an example, a module may refer to three questions which have to be asked in sequence during an outbound interview, each detected by a single atom. Like atoms, modules are declared for a specific speech-to-text service. In this regard, it is worth noting that a module should be homogeneous, that is, it should only contain atoms pertaining to the same service.

Finally, modules are used in the context of an inbound or outbound service offered by the contact center (e.g., the toll-free number of an airline company). 

\subsection{Data description}
\label{datatotag}

In this section, we give a short account of the actual data for 
the considered outbound survey service. 
The list of defined keywords is the following:
\begin{itemize}
\item \emph{age}: the agent asked the interviewed person his/her age;
%\item \emph{badwords}: the agent said some unwanted words, which may be precisely stated in the terms-of-service, or some general bad words;
\item \emph{call\_permission}: the agent asked the called person for the permission to conduct the survey;
\item \emph{duration\_info}: the agent informed the called person about the duration of the survey;
\item \emph{family\_unit}: the agent asked the called person about his/her family unit; 
\item \emph{greeting\_initial}: the agent introduced himself/herself correctly at the beginning of the phone call;
\item \emph{greeting\_final}: the agent pronounced the scripted  goodbye phrases;
\item \emph{person\_identity}: the agent asked the called person for a confirmation of his/her identity;
\item \emph{privacy}: the agent informed the called person about the privacy implications of the phone call;
\item \emph{profession}: the agent asked the interviewed person about his/her job;
\item \emph{question\_1}: the agent asked the first question of the survey;
\item \emph{question\_2}: the agent asked the second question of the survey;
\item \emph{question\_3}: the agent asked the third question of the survey.
\end{itemize}

Table \ref{tab:esempifrasi} shows some hand-made transcriptions with the associated keywords. Once again, observe that keywords are general concepts, which are independent from the specific service under consideration.

\begin{table*}[t]
\centering
\caption{Some exemplary, hand-made transcriptions with the associated keywords. Data has been anonymized, and punctuation has been added to the transcriptions for ease of reading}
\label{tab:esempifrasi}
\resizebox{\textwidth}{!}{
\begin{tabular}{p{6.5cm}p{6.5cm}p{3cm}}
\hline
\textbf{Phrase (It)}                                                                                        & \textbf{Phrase (En)}                                                                                & \textbf{Tags}   \\
\hline 
Si pronto buongiorno sono X dalla X di X, parlo con la signora X?                                   & Hello, my name is X and I am calling from X of X, am I talking with Mrs X?                  & greeting\_initial, \mbox{person\_identity} \\
\hline
Lei è pensionato. Ultima domanda, senta, a livello statistico la data solo di nascita... millenovecento...? & You are a pensioner. Last question, listen, statistically, the birth date only... nineteen hundred...? & age, profession                     \\
\hline
Ho capito. Posso chiederle il nome di battesimo?                                                            & Understood. May I ask you for your first name?                                               & person\_identity                    \\
\hline
Mi permette? Trenta secondi, tre domande velocissime...                                                     & May I? Thirty seconds, three quick questions...                                                     & duration\_info  \\
\hline
\end{tabular}
}
\end{table*}

To allow for the training of the machine learning models, and for the evaluation of all the approaches, a set of 4884 text segments originated from 482 distinct outbound sessions of the considered survey service have been manually tagged by domain experts,
%with all the relevant keywords,
 so that each instance is characterized by the transcription, and by a list of Boolean attributes that track the presence or absence of each specific keyword. Each session has been independently transcribed by Kaldi and Google, in order to evaluate the tagging performance based on both services. The transcriptions may indeed contain some errors, and this is deliberate, since we are not interested in assessing the tagging performances over perfectly transcribed data, but rather in assessing the noise tolerance of the tagging models.
 
Table \ref{tab:numerotags} reports, for each tag, the number of instances in which it is present or not. The resulting dataset has then been split into a training ($75\%$, 3696 instances) and a test ($25\%$, 1188 instances) set, according to a \emph{stratified random sampling by group} approach, where each single session is a group on its own. This allowed us not to fragment segments belonging to a single session between the two sets and, moreover, to preserve the keyword distribution between them.

\begin{table}[t]
\centering
\caption{Number and percentage of instances in which each tag is present (P) or not (NP), in the full dataset}
\label{tab:numerotags}
%\resizebox{\linewidth}{!}{
\begin{tabular}{lllll}
\hline
\textbf{Keyword name} & \textbf{\# P} & \textbf{\% P} & \textbf{\# NP} & \textbf{\% NP} \\
\hline
age                   & 638                  & 13.1                 & 4246                 & 86.9                 \\
call\_permission      & 565                  & 11.6                 & 4319                 & 88.4                 \\
duration\_info        & 491                  & 10.1                 & 4393                 & 89.9                 \\
family\_unit          & 506                  & 10.4                 & 4378                 & 89.6                 \\
greeting\_initial     & 560                  & 11.5                 & 4324                 & 88.5                 \\
greeting\_final       & 453                  & 9.3                  & 4431                 & 90.7                 \\
person\_identity      & 600                  & 12.3                 & 4284                 & 87.7                 \\
privacy               & 440                  & 9.0                  & 4444                 & 91.0                 \\
profession            & 391                  & 8.0                  & 4493                 & 92.0                 \\
question\_1           & 516                  & 10.6                 & 4368                 & 89.4                 \\
question\_2           & 496                  & 10.2                 & 4388                 & 89.8                 \\
question\_3           & 500                  & 10.2                 & 4384                 & 89.8         \\
\hline
\end{tabular}
%}
\end{table}

\subsection{Tagging with Regular Expressions}
\label{sec:taggingregex}
Domain expert-defined regular expressions (\emph{RegEx})  have already been used 
%in the past for the purpose of extracting 
in order to extract knowledge from textual data, e.g., \citep{gold2008extracting}.

In the present work, 12 \emph{RegEx} atoms, which encapsulate the semantic content of utterance transcriptions, have been manually defined, one for each of the keywords that have to be recognized. By exploiting regular expressions, these atoms make it possible to deal with the plain text of the transcriptions, without requiring any further data transformation process, like, for instance, stemming or stopwords removal. As the defined regular expressions try to locate linguistic patterns inside transcriptions, we refer to this tagging strategy as a  {\em linguistic} approach. 

In order to make it evident how difficult is the process of defining suitable regular expressions, Table \ref{tab:complregex} provides some complexity measures on them. As it can be observed, expressions are typically quite long when measured in terms of the number of characters and, most notably, they are inherently complex, as the corresponding Deterministic Finite Automata (DFA) have a large number of states. 
As an example, the regular expression \emph{.*(you born\textbar age\textbar how old are you\textbar you were born in\textbar (in\textbar from) which year\textbar less than eighty).*}, translated and reworked from Italian into English, recognizes the atom \emph{age}.

\begin{table}[t]
\centering
\caption{Complexity measures on the hand-made regular expressions}
\label{tab:complregex}
%\resizebox{\linewidth}{!}{
\begin{tabular}{lll}
\hline
\textbf{Keyword name} & \textbf{\# regex chars} & \textbf{\# DFA states} \\
\hline
age                   & 94                         & 79                        \\
call\_permission      & 143                        & 102                       \\
duration\_info        & 158                        & 104                       \\
family\_unit          & 135                        & 97                        \\
greeting\_initial     & 30                         & 23                        \\
greeting\_final       & 143                        & 99                        \\
person\_identity      & 189                        & 121                       \\
privacy               & 101                        & 65                        \\
profession            & 103                        & 85                        \\
question\_1           & 107                        & 80                        \\
question\_2           & 244                        & 126                       \\
question\_3           & 209                        & 141       \\
\hline
\end{tabular}
%}
\end{table}

\subsubsection{Evaluation of tagging performance}
\label{sec:regexeval}

Each of the 1188 segments in the Kaldi and Google test sets has been evaluated against the presence of every possible keyword, looking for the concordance between its manual annotations and the ones given by the regular expressions. This is, to all intents and purposes, a {\em supervised classification} problem, and the performances of our model, shown in Table \ref{tabellaregextag}, can be measured by standard metrics:
\begin{itemize}
\item \emph{accuracy}, which tracks the fraction of times when a tag has been correctly identified in a segment as present or absent;
\item \emph{precision}, which is the fraction of segments in which a specific tag has been identified as present by the method, and in which the tag is indeed present;
\item \emph{recall}, that reports the proportion of segments presenting the specific tag, that have been in fact identified as such by the method; 
\item \emph{true negative rate} (TNR), which shows the proportion of segments not presenting the specific tag, that have been classified as negative by the method; 
\end{itemize}

The above-described metrics will be used throughout the paper to test the performance of the different approaches to tagging. As shown by Table \ref{tabellaregextag}, the regular expressions are capable of reaching a very high accuracy in tagging the segments, considering both Kaldi and Google transcriptions. Such a result, however, is not indicative of the true performance, since it is biased by the large disproportion between the instances in which the tags are present and those in which they are not. More useful indicators are the true negative rate, the precision and, especially, the recall, which reveals the tags that are most difficult to identify. As a matter of fact, looking at Kaldi transcribed data, satisfactory results are obtained for most of the tags, except for \emph{call\_permission}, that has a fairly low precision and recall. Then, also the tags \emph{greeting\_initial}, \emph{person\_identity}, and \emph{profession} show a recall performance somewhat under the average, and seem in general hard to identify. As for the results on Google transcriptions, accuracy tends to be slightly lower than Kaldi, with the notable exception of \emph{call\_permission}. Moreover, while precision seems to be better in general, recall is overall significantly lower. While this may come as a surprise, given the fact that Google transcripts are in general less error prone than Kaldi ones, it can, however, be easily explained, since it turns out that domain experts have defined the regular expressions based on Kaldi transcriptions only. This highlights the necessity of dealing with Kaldi and Google transcriptions separately, an approach that we are indeed going to pursue in the following sections.

We conclude by remarking that the definition of suitable regular expressions comes with a cost. As already discussed, each keyword is evaluated by a corresponding regular expression, which must be defined by domain experts on the basis of their personal knowledge, through a delicate, empirical, and long refining process.

\begin{table*}[t]
\centering
\caption{Tagging performance with the linguistic approach over Kaldi (K) and Google (G) transcriptions}
\label{tabellaregextag}
\begin{tabular}{lcccccccc}
\hline
& \multicolumn{2}{c}{\textbf{Accuracy}} 
& \multicolumn{2}{c}{\textbf{Precision}} 
& \multicolumn{2}{c}{\textbf{Recall}} 
& \multicolumn{2}{c}{\textbf{TNR}} \\ 
\textbf{Keyword} & \textbf{K} & \textbf{G} & \textbf{K} & \textbf{G} & \textbf{K} & \textbf{G} & \textbf{K} & \textbf{G}  \\
\hline
age                   & 0.971 &  0.902    & 0.906 &  0.988     & 0.884  &  0.435 & 0.985  & 0.999  \\
call\_permission      & 0.918 &  0.949    & 0.697 &  0.732     & 0.539  &  0.598 & 0.969  & 0.981    \\
duration\_info        & 0.982 &  0.942    & 0.884 &  0.635     & 0.953  &  0.922 & 0.985  & 0.944    \\
family\_unit          & 0.969 &  0.944    & 0.857 &  0.897     & 0.864  &  0.684 & 0.982  & 0.987    \\
greeting\_initial     & 0.949 &  0.919    & 0.987 &  0.988     & 0.549  &  0.488 & 0.999  & 0.999    \\
greeting\_final       & 0.991 &  0.977    & 0.991 &  1.000     & 0.916  & 0.747  & 0.999  & 1.000    \\
person\_identity      & 0.944 &  0.934    & 0.818 & 0.925      & 0.657  & 0.609  & 0.981  & 0.991    \\
privacy               & 0.991 &  0.908    & 0.963 & 1.000      & 0.937  & 0.030  & 0.996  & 1.000    \\
profession            & 0.970 &  0.908    & 0.969 & 1.000      & 0.646  & 0.101  & 0.998  & 1.000    \\
question\_1           & 0.967 &  0.987    & 0.990 & 1.000      & 0.723  & 0.873  & 0.999  & 1.000    \\
question\_2           & 0.978 &  0.964    & 0.943 & 0.990      & 0.833  & 0.728  & 0.994  & 0.999    \\
question\_3           & 0.960 &  0.964    & 0.942 & 0.976      & 0.653  & 0.690  & 0.995  & 0.998   \\
\hline
%\textbf{Average} & 0.9656                 & 0.9123                  & 0.7619               & 0.9903            \\ 
%\hline
\end{tabular}
\end{table*}

\subsection{Tagging with Logistic Regression}
\label{sec:tagginglogistic}

Let us consider now an alternative approach to tagging based on machine learning techniques. One of its advantages is that it allows us to get rid of the support from domain experts in the definition of suitable rules for the identification of tags in the transcriptions. 

Since we are interested in determining the presence or absence of a tag in a segment, we focus on a subset of machine learning algorithms, namely, classification algorithms. Moreover, given that the  classification task under consideration is a binary one, and we are interested not only in getting a result, but also in somehow tracking the decision process that has been followed,
%(which can be studied and justified by domain experts)
 we opt for {\em logistic regression}.

Logistic regression generates an interpretable model, as it exposes the relative importance of the features and how they contribute to the overall outcome. It should be remarked that interpretability is a fundamental requirement for the company, since it allows for the validation of the model by domain experts, that, in turn, constitutes an essential step before entering the production phase. This, together with assessments regarding the complexity of the problem, led us to this solution, rather than, for instance, to the use of support vector machines, Conditional Random Fields, or neural networks.

\subsubsection{Data preparation}
\label{sec:datapreparationlogistic}

Starting from the data described in Section \ref{datatotag}, the
training and test datasets have been prepared independently considering Kaldi and Google transcripts as follows. First of all, the text has been converted to lowercase, and all non-alphabetical characters have been removed from each segment. Next, all Italian stopwords have been also removed. Then, \emph{Porter2} stemming algorithm for Italian has been applied. Finally, we extracted from the processed text all unigrams, bigrams, and trigrams with a frequency of at least $1\%$, as calculated on the training dataset. All operations have been carried out by means of 
custom-made \emph{Python} scripts, by making use of the \emph{NLTK} library.

Three training and test dataset pairs have then been created for each keyword, considering the different  ngram lengths. Each dataset consists of a set of predictor attributes that track whether the corresponding ngram is present or not in the transcription, plus the class. As we have already mentioned, such a setting has been replicated for both Kaldi and Google transcriptions, so to train two independent sets of models, following the observation of Section \ref{sec:regexeval}. Thus, given the 12 possible tags, the 3 different ngram lengths, and the 2 transcription models that have been considered, we generated a total of 72 training and test dataset pairs. For each dataset, Table \ref{tab:beforeattrselection} shows the number of originally considered predictors. As it can be seen, the number of attributes generated from the Google transcriptions is considerably higher than Kaldi's. This might be explained by the small dictionary employed by the latter transcription model.

\begin{table}[t!]
\centering
\caption{Original number of attributes per Unigram (Uni), Bigram (Bi) and Trigram (Tri) datasets, considering Kaldi and Google transcriptions}
\label{tab:beforeattrselection}
%\resizebox{\linewidth}{!}{
\begin{tabular}{lccc}
\hline
& \multicolumn{3}{c}{\textbf{Number of attributes}} \\
\textbf{Transcription model} &  \textbf{Uni} &  \textbf{Bi} &  \textbf{Tri}  \\
\hline
Kaldi & 177 & 154 & 102 \\
Google & 444 & 747 & 664 \\
\hline
\end{tabular}
%}
\end{table}

\subsubsection{Training of the models}
\label{sec:logistictraining}

In order to train the models, we relied on WEKA's \emph{Logistic} algorithm \citep{witten2016data}. However, given the very high number of predictors involved, an attribute selection step has been applied in order to reduce the dimension of the datasets before actually proceeding with the training phase.

The design of a feature selection step entails the selection of a search strategy, to guide the incremental generation of the feature set, and of an evaluation strategy, to score the candidates (subsets of attributes). 

As for the search strategy, we considered WEKA's \emph{BestFirst} \citep{heuristics1984intelligent}, which is an implementation of \emph{beam search}, that searches the space of attribute subsets by greedy hill climbing, augmented with a backtracking capability. 

As for the evaluation strategy, WEKA's multivariate filter \emph{CfsSubsetEval} \citep{hall1999correlation} has been used. Such a filter evaluates the worthiness of a subset of attributes by considering the individual predictive power of each feature, together with the degree of redundancy between them, preferring subsets of features that are highly correlated with the class while having low intercorrelation among them.

The results of the attribute selection phase are reported in Table \ref{tab:attrselection}. As it turned out, many of the attributes have been discarded, meaning, perhaps not surprisingly, that only certain ngrams are useful for determining the presence or absence of each specific keyword. Once again, observe that the Google based datasets have a higher number of attributes than Kaldi's.

\begin{table}[t!]
\centering
\caption{Number of attributes per Unigram (Uni), Bigram (Bi) and Trigram (Tri) datasets, after the attribute selection step, considering Kaldi (K) and Google (G) transcriptions}
\label{tab:attrselection}
%\resizebox{\linewidth}{!}{
\begin{tabular}{lcccccc}
\hline
& \multicolumn{2}{c}{\textbf{Uni}} 
& \multicolumn{2}{c}{\textbf{Bi}} 
& \multicolumn{2}{c}{\textbf{Tri}} \\
\textbf{Keyword} &  \textbf{K} & \textbf{G} & \textbf{K} & \textbf{G} & \textbf{K} & \textbf{G}   \\
\hline
age                   & 8    &  45                        & 11    &  66                      & 8    &       68                   \\
call\_permission      & 8    &  17                        & 9     &  21                     & 9    &        71                  \\
duration\_info        & 14   &  18                        & 11    &  22                      & 10   &       64                   \\
family\_unit          & 15   &  20                        & 12    &  65                      & 10   &       58                   \\
greeting\_initial     & 11   &  18                        & 14    &  69                      & 13   &       72                   \\
greeting\_final       & 11   &  14                        & 13    &  17                      & 11   &       58                   \\
person\_identity      & 8    &  17                        & 9     &  72                      & 9    &   80                       \\
privacy               & 10   &  13                        & 12    &  70                      & 10   &       69                   \\
profession            & 11   &  22                        & 9     &  22                      & 24   &       66                   \\
question\_1           & 11   &  15                        & 13    &  74                      & 16   &    72                      \\
question\_2           & 10   &  14                        & 10    &  66                      & 9    &       68                   \\
question\_3           & 12   &  14                        & 10    &  58                      & 12   &       62     \\
\hline
\end{tabular}
%}
\end{table}

\subsubsection{Evaluation of tagging performance}
\label{sec:logisticeval}

We considered the performance of the models based on, respectively, unigrams, bigrams, and trigrams on each of the 1188 segments in the Kaldi and Google test sets. The results are reported in Table \ref{tabellalogregtag}. 

Considering both Kaldi and Google transcripts and models, precision does not show a consistent behaviour among the ngram sizes, while in many cases recall decreases sharply with the larger ngram sizes. It is also interesting to observe that, as the ngrams grow in size, the accuracy seems to slightly decrease. This behavior can be explained in two ways: $(i)$ there might not be a sufficient number of instances in the datasets to justify the adoption of bigrams and trigrams; $(ii)$ the presence or absence of a keyword in calls generated in the context of the specific service that has been considered is not a difficult concept to learn, and thus unigrams are sufficient for the task.

\begin{table*}[t!]
\centering
\caption{Tagging performance with the Logistic Regression approach over Kaldi (K) and Google (G) transcriptions}
\label{tabellalogregtag}
\begin{tabular}{lcccccccc}
\hline
& \multicolumn{2}{c}{\textbf{Accuracy}} 
& \multicolumn{2}{c}{\textbf{Precision}} 
& \multicolumn{2}{c}{\textbf{Recall}} 
& \multicolumn{2}{c}{\textbf{TNR}} \\ 
\textbf{Keyword} & \textbf{K} & \textbf{G} & \textbf{K} & \textbf{G} & \textbf{K} & \textbf{G} & \textbf{K} & \textbf{G}  \\
\hline
age-uni               & 0.966 & 0.953     & 0.897 & 0.957      & 0.854 & 0.988   & 0.984 & 0.788     \\
age-bi                & 0.961 & 0.953     & 0.940 & 0.968      & 0.768 & 0.976   & 0.992 & 0.842     \\
age-tri               & 0.924 & 0.934     & 0.963 & 0.940      & 0.470 & 0.982   & 0.997 & 0.701 \\
\hline
call\_permission-uni  & 0.931 & 0.959     & 0.798 & 0.765      & 0.560 & 0.713   & 0.981 & 0.981  \\
call\_permission-bi   & 0.910 & 0.947     & 0.713 & 0.727      & 0.404 & 0.552   & 0.978 & 0.982 \\
call\_permission-tri  & 0.902 & 0.942     & 0.705 & 0.736      & 0.305 & 0.448   & 0.983 & 0.986 \\
\hline
duration\_info-uni    & 0.967 & 0.955     & 0.908 & 0.807      & 0.773 &  0.696  & 0.991 & 0.982      \\
duration\_info-bi     & 0.963 & 0.951     & 0.912 & 0.821      & 0.727 &  0.628  & 0.992 & 0.986      \\
duration\_info-tri    & 0.913 & 0.939     & 0.833 & 0.785      & 0.234 &  0.500  & 0.994 & 0.986 \\
\hline
family\_unit-uni      & 0.963 & 0.972     & 0.893 & 0.930      & 0.758 &  0.868  & 0.989 & 0.989      \\
family\_unit-bi       & 0.955 & 0.943     & 0.944 & 0.960      & 0.636 &  0.625  & 0.995 & 0.996 \\
family\_unit-tri      & 0.900 & 0.915     & 1.000 & 0.918      & 0.099 &  0.441  & 1.000 & 0.993  \\
\hline
greeting\_initial-uni & 0.977 & 0.949     & 0.921 &  0.900     & 0.872 &  0.759  & 0.991 & 0.985      \\
greeting\_initial-bi  & 0.961 & 0.963     & 0.922 &  0.957     & 0.714 &  0.795  & 0.992 & 0.993      \\
greeting\_initial-tri & 0.931 & 0.940     & 0.947 &  0.964     & 0.406 &  0.639  & 0.997 & 0.996 \\
\hline
greeting\_final-uni   & 0.994 & 0.997     & 0.991 &  0.990     & 0.950 &  0.980  & 0.999 & 0.999  \\
greeting\_final-bi    & 0.990 & 0.989     & 0.965 &  0.968     & 0.933 &  0.909  & 0.996 & 0.997 \\
greeting\_final-tri   & 0.986 & 0.985     & 0.964 &  1.000     & 0.891 &  0.838  & 0.996 & 1.000 \\
\hline
person\_identity-uni  & 0.937 & 0.943     & 0.750 &  0.868     & 0.679 &  0.733  & 0.971 & 0.980     \\
person\_identity-bi   & 0.932 & 0.935     & 0.811 &  0.859     & 0.533 &  0.683  & 0.984 & 0.980 \\
person\_identity-tri  & 0.910 & 0.923     & 0.800 &  0.891     & 0.292 &  0.559  & 0.991 & 0.988 \\
\hline
privacy-uni           & 0.991 & 0.998     & 0.955 &  0.990     & 0.946 &  0.990  & 0.995 & 0.999      \\
privacy-bi            & 0.988 & 0.995     & 0.945 &  0.990     & 0.928 &  0.960  & 0.994 & 0.999 \\
privacy-tri           & 0.973 & 0.979     & 0.891 &  0.976     & 0.811 &  0.792  & 0.990 & 0.998 \\
\hline
profession-uni        & 0.987 & 0.975     & 0.893 &  0.873     & 0.958 &  0.881  & 0.990 & 0.985      \\
profession-bi         & 0.929 & 0.944     & 1.000 &  0.889     & 0.125 &  0.514  & 1.000 & 0.993  \\
profession-tri        & 0.919 & 0.911     & -     &  0.769     & 0.000 &  0.184  & 1.000 & 0.994  \\
\hline
question\_1-uni       & 0.984 & 0.993     & 0.968 &  0.972     & 0.891 &  0.955  & 0.996 & 0.997      \\
question\_1-bi        & 0.985 & 0.995     & 0.984 &  0.982     & 0.883 &  0.973  & 0.998 & 0.998 \\
question\_1-tri       & 0.980 & 0.987     & 0.991 &  1.000     & 0.832 &  0.873  & 0.999 & 1.000  \\
\hline
question\_2-uni       & 0.984 & 0.990     & 0.924 &  0.970     & 0.917 & 0.949   & 0.992 & 0.996      \\
question\_2-bi        & 0.976 & 0.991     & 0.933 &  0.992     & 0.817 & 0.934   & 0.993 & 0.999 \\
question\_2-tri       & 0.974 & 0.977     & 0.932 &  0.959     & 0.800 & 0.853   & 0.993 & 0.995 \\
\hline
question\_3-uni       & 0.985 & 0.990     & 0.942 & 0.973      & 0.911 & 0.931   & 0.993 & 0.997      \\
question\_3-bi        & 0.976 & 0.987     & 0.936 & 0.964      & 0.823 & 0.914   & 0.993 & 0.996 \\
question\_3-tri       & 0.973 & 0.985     & 0.942 & 0.981      & 0.790 & 0.879   & 0.994 & 0.998   \\
\hline
\end{tabular}
\end{table*}

\subsection{Tagging with a hybrid approach}
\label{sec:hybridapproach}

In this section, we explore a simple hybrid approach to tagging that combines the linguistic and machine learning based strategies.

In developing it, we took into account that, even though the company is definitely interested in finding patterns of problematic calls, it is also true that the supervising staff ought not to be overwhelmed with false notifications (a correct call signaled as problematic) carrying requests for verification. 

On the basis of empirical evaluations, it has been found out that a simple combination of the results of the linguistic and machine learning approaches leads to satisfactory results: a keyword is considered to be present if at least one of the two individual strategies marks it as present.

Table \ref{tab:hybridperf} reports the detailed figures, obtained from the combination between the results given by the regular expressions and the unigram logistic models. As it can be observed for both Kaldi and Google transcripts, the accuracies typically show a slight improvement over the two individual approaches. Moreover, as expected, given the specific combination strategy, the precision has decreased a little bit, while the recall has gained a considerable boost. This is exactly what had to be accomplished: it is now more likely to predict a tag as present, and thus the problematic calls are limited to the most serious cases.

\begin{table*}[t]
\centering
\caption{Tagging performance with the hybrid approach, over Kaldi (K) and Google (G) transcriptions}
\label{tab:hybridperf}
\begin{tabular}{lcccccccc}
\hline
& \multicolumn{2}{c}{\textbf{Accuracy}} 
& \multicolumn{2}{c}{\textbf{Precision}} 
& \multicolumn{2}{c}{\textbf{Recall}} 
& \multicolumn{2}{c}{\textbf{TNR}} \\ 
\textbf{Keyword} & \textbf{K} & \textbf{G} & \textbf{K} & \textbf{G} & \textbf{K} & \textbf{G} & \textbf{K} & \textbf{G}  \\
\hline
age                   & 0.966 &  0.962    & 0.897 &   0.943    & 0.854 &  0.836  & 0.984 &  0.989    \\
call\_permission      & 0.923 &  0.955    & 0.699 &   0.723    & 0.610 &  0.737  & 0.965 &  0.976    \\
duration\_info        & 0.982 &  0.941    & 0.879 &   0.637    & 0.961 &  0.958  & 0.984 &  0.939    \\
family\_unit          & 0.962 &  0.968    & 0.800 &   0.880    & 0.879 &  0.922  & 0.973 &  0.977    \\
greeting\_initial     & 0.981 &  0.942    & 0.917 &   0.900    & 0.910 &  0.721  & 0.990 &  0.985    \\
greeting\_final       & 0.995 & 0.995     & 0.991 &   0.983    & 0.958 &  0.964  & 0.999 & 0.999     \\
person\_identity      & 0.946 & 0.949     & 0.784 &   0.846    & 0.803 &  0.796  & 0.965 & 0.976     \\
privacy               & 0.992 & 0.999     & 0.955 &   0.993    & 0.955 &  1.000  & 0.995 & 0.999     \\
profession            & 0.987 & 0.981     & 0.886 &   0.925    & 0.969 &  0.905  & 0.989 & 0.991     \\
question\_1           & 0.985 & 0.993     & 0.969 &   0.953    & 0.898 &  0.980  & 0.996 & 0.994     \\
question\_2           & 0.984 & 0.992     & 0.924 &   0.963    & 0.917 &  0.970  & 0.992 & 0.995     \\
question\_3           & 0.985 & 0.993     & 0.934 &   0.978    & 0.919 &  0.958  & 0.993 & 0.997    \\
\hline
%\textbf{Average}              & 0.9739                 & 0.8863                  & 0.8860               & 0.9853     \\
%\hline
\end{tabular}
\end{table*}

\subsection{A critical evaluation of the proposed solutions}
\label{taggingeval}

We conclude the section with a comparison of the proposed solutions.
Consider Table \ref{tab:avgperf}. As it turns out, the regular expression approach has an overall accuracy comparable to the one exhibited by the unigram-based machine learning approach, which has already been shown to be the best ngram choice for the problem. However, the unigram approach exhibits a clearly better behaviour for what concerns the recall which, as we already pointed out, is the key performance indicator for the considered problem. In addition, the definition of suitable regular expressions is a tedious and error-prone process, that has to be done manually, which in turn consumes a significant amount of company time and personnel resources. As for the hybrid approach, it has proven itself capable of boosting recall, although at the cost of a slight decrease in precision. Finally, it should be observed that the results based on Kaldi transcripts are just slightly inferior to those obtained on Google ones, especially when considering the hybrid solution. Thus, at least for the considered tagging task, the in-house and low-cost developed solution seems to be sufficient to accomplish the intended goals.

\begin{table*}[t]
\centering
\caption{Average performance per method, over Kaldi (K) and Google (G) transcriptions}
\label{tab:avgperf}
\begin{tabular}{lcccccccc}
\hline
& \multicolumn{2}{c}{\textbf{Accuracy}} 
& \multicolumn{2}{c}{\textbf{Precision}} 
& \multicolumn{2}{c}{\textbf{Recall}} 
& \multicolumn{2}{c}{\textbf{TNR}} \\ 
\textbf{Keyword} & \textbf{K} & \textbf{G} & \textbf{K} & \textbf{G} & \textbf{K} & \textbf{G} & \textbf{K} & \textbf{G}  \\
\hline
Regular expressions & 0.966 & 0.942          & 0.912  &   0.928        & 0.763  &  0.575      & 0.990  &   0.992       \\
Logistic, unigram   & 0.972 & 0.973          & 0.903  &   0.916        & 0.839  &  0.870      & 0.989  &   0.973       \\
Logistic, bigram    & 0.961 & 0.966          & 0.917  &   0.923        & 0.691  &  0.789      & 0.992  &   0.980          \\
Logistic, trigram   & 0.940 & 0.951          & -      &   0.910        & 0.494  &  0.666      & 0.995  &   0.895  \\
Hybrid              & 0.974 & 0.973          & 0.886  &   0.894        & 0.886  &  0.896      & 0.985  &   0.985  \\
\hline
\end{tabular}
\end{table*}

\iffalse
\begin{table}[b]
\centering
\caption{Average performance per method.}
\label{tab:avgperf}
\begin{tabular}{lllll}
\hline
\textbf{Method}     & \textbf{Avg accuracy} & \textbf{Avg precision} & \textbf{Avg recall} & \textbf{Avg TNR} \\
\hline
Regular expressions & 0.9656                 & 0.9123                  & 0.7619               & 0.9903            \\
Logistic, unigram   & 0.9722                 & 0.9033                  & 0.8390               & 0.9893            \\
Logistic, bigram    & 0.9605                 & 0.9171                  & 0.6909               & 0.9924            \\
Logistic, trigram   & 0.9404                 & -                       & 0.4941               & 0.9946            \\
Hybrid              & 0.9739                 & 0.8863                  & 0.8860               & 0.9853     \\
\hline
\end{tabular}
\end{table}
\fi

\section{Tagging of segments: an alternative approach}
%J48S and its application to the tagging of chunks}
\label{sec:decisiontree}

In the light of the baseline results obtained with the traditional approaches presented in the previous section, here we illustrate an alternative solution to the tagging of segments that relies on J48S decision tree induction algorithm (for a detailed account of the algorithm, we refer the reader to~\citep{j48s}), based on J48 (Weka's implementation of C4.5 Release 8 \citep{witten2016data}) and VGEN sequential pattern extraction algorithm  \citep{fournier2014vgen}. The distinctive feature of J48S is that it is able of mixing the usage of sequential, such as text, and classical, that is, categorical or numeric, data on the same execution cycle.

\subsection{The decision tree induction algorithm J48S}

A J48S decision tree is built recursively, starting from the root, following the traditional TDIDT (\emph{Top Down Induction of Decision Trees}) approach: at each node, the attribute that best partitions the training data, according to a predefined score, is chosen as a test to guide the partitioning of instances into child nodes, and the process continues until a sufficiently high degree of purity (with respect to the target class), or a minimum cardinality constraint (with respect to the number of instances reaching the node), is achieved in the generated partitions. %

As for the splitting process, the attribute with the highest \emph{normalized gain} (NG) \citep{jun1997new} is chosen. To this end, the algorithm first evaluates all numeric and categorical attributes according to their NG; then, by exploiting a specifically adapted version of VGEN algorithm,
it extracts a single pattern from each sequential attribute, 
%again 
evaluating each of them on the basis of its NG. 

To counter possible pattern overfitting effects, two measures are employed:
\begin{itemize}
\item \emph{generator patterns} are extracted, where a pattern is considered to be a generator if there exists no other pattern which is contained in it and has the same frequency. According to the MDL (\emph{Minimum Description Length}) principle \citep{rissanen1978modeling}, it can be argued that generators are preferable over classic patterns for model selection and classification, since they may be less prone to overfitting effects \citep{lo2008mining};
\item the pattern extracted by VGEN is not the most informative one, but the one minimizing the following formula, again inspired by the MDL principle:
\begin{align*}
W*(1-pattern_{rNG}) + (1-W)*pattern_{rlen}
\end{align*}
where $W\in[0,1]$ is a user-customizable weight,  $pattern_{rNG}\in[0,1]$ is the relative normalized gain of the pattern with respect to the highest normalized gain observed among the patterns, and $pattern_{rlen}\in[0,1]$ is the relative length of the pattern, with respect to the longest pattern that has been discovered (in terms of the number of items). Intuitively, the larger the user sets the weight, the longer and more accurate (on the training set) the extracted pattern is. On the contrary, selecting a low weight should lead to a shorter, and more general, pattern being selected. In essence, parameter $W$ configures itself as a way of controlling the abstraction level of the extracted patterns.
\end{itemize}

Table \ref{attrsj48s} lists all the parameters that have been added to J48S, with respect to plain J48. There are just 6 of them, each with a rather intuitive meaning.

\begin{figure}[t]
\centering
\includegraphics[scale=0.9]{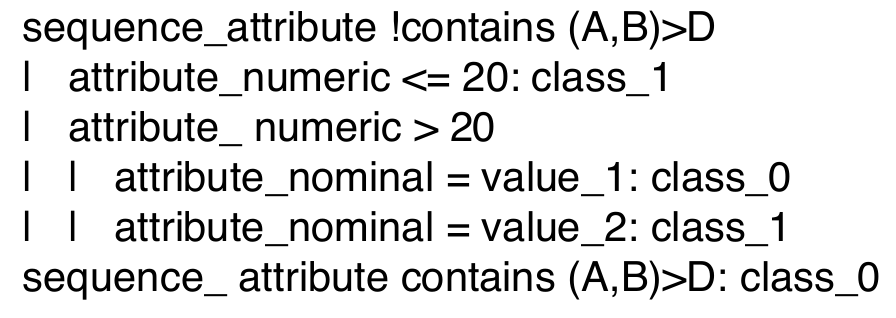}
\caption{A typical J48S decision tree built by means of WEKA.}\label{dectreeimage}
\end{figure}

In Figure \ref{dectreeimage}, a simple example of a J48S decision tree is given, built by integrating the extension to WEKA data mining suite \citep{witten2016data}. The reference dataset is characterized by three features:  \emph{sequence\_attribute}, which is of type string and represents a sequence of itemsets, \emph{attribute\_numeric}, which has an integer value, and  \emph{attribute\_nominal}, which may only take one out of a predefined set of values. The class is binary, and has two labels: \emph{class\_0} and \emph{class\_1}. The first test made by the tree is whether the given sequence contains the pattern $(A,B)>D$ or not, meaning that there should be an itemset containing $A$, $B$, followed (within the maximum gap constraint) by an itemset containing $D$. If this is the case, then the instance is labeled with \emph{class\_0}; otherwise, the tree proceeds by testing on the numeric and nominal attributes.

\begin{table*}[t]
\centering
\caption{Custom parameters in J48S}
\label{attrsj48s}
\begin{tabular}{lll}
\hline
\textbf{Parameter name} & \textbf{Default} & \textbf{Description}                                              \\
\hline
maxGap                  & 2                      & maximum gap allowance between itemsets (1 = contiguous)           \\
maxPatternLength        & 20                     & maximum length of a pattern (number of items)                   \\
maxTime                 & 30                     & maximum allowed running time of the algorithm per call           \\
minSupport              & 0.5                    & minimum support of a pattern                                      \\
patternWeight           & 0.5                    & weight used in VGEN algorithm for the extraction of the result         \\
useIGPruning            & True                   & toggle information gain pruning of the pattern search space \\
\hline 
\end{tabular}
\end{table*}

\subsection{Tagging with J48S}
\label{sec:tagj48s}

In the following, we present the setup and the results of the experimentation carried out with J48S, with the aim of tagging call transcripts.

\subsubsection{Data preparation}
\label{sec:datapreparationdecisiontree}

Starting from the data described in Section \ref{datatotag},  the training and test datasets have been prepared as follows. As done for the logistic regression case, the text has been converted to lowercase, and all non-alphabetical characters have been removed from each segment. Then, stopwords have been discarded, and \emph{Porter2} stemming algorithm has been applied. A training and a test dataset have been created for each keyword and transcription model, where each of the instances is characterized by just two attributes: the processed text of the segment, and an attribute identifying the presence or absence of the specific keyword.

\subsubsection{Training of the model}
\label{sec:decisiontreetraining}

J48S has been implemented within the WEKA data mining suite \citep{witten2016data}.  
All J48S parameters have been left with their default values, except for \emph{minSupport}, which has been set to 0.01 (1\%), in order to allow for the extraction of uncommon patterns, without performing any specific kind of tuning.%, and \emph{patternWeight}, which has been set to 1, considering that the survey service requires very specific phrases to be pronounced by the agents.

\subsubsection{Evaluation of tagging performance}
\label{sec:decisiontreeeval}

Following the same approach as the one used for the evaluation of the \emph{RegEx} (see Section \ref{sec:regexeval}) and logistic regression (see Section \ref{sec:logisticeval}) strategies, we measured the performance of J48S in determining the presence or absence of each keyword in the same test set made of 1188 segments. As shown by the results reported in Table \ref{tabellatreetag}, J48S is capable of matching the accuracy of the hybrid approach. Moreover, though having a slightly worse precision than regular expression and logistic models, it surpasses the hybrid method score. As for the recall, it is only beaten by the hybrid solution, which nonetheless is much more complex, making use of two distinct strategies.

\begin{figure}[t]
\centering
\includegraphics[scale=0.9]{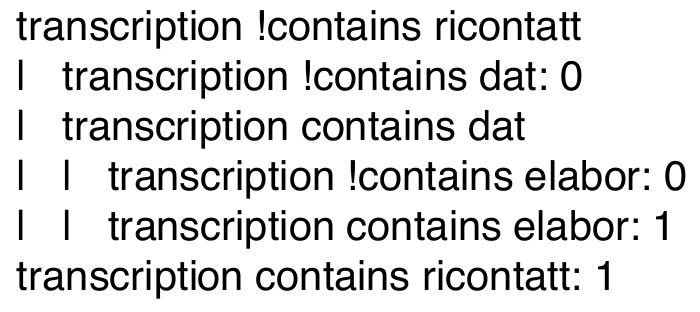}
\caption{J48S decision tree recognizing the presence of the tag \emph{greeting\_final}.}\label{alberosalutifinali}
\end{figure}

All in all, we can conclude that J48S is able of providing results comparable to the ones of the hybrid approach.
Moreover, it is worth mentioning that: $(i)$ the measured performances have been obtained training the algorithm with the default parameters (except for the minimum support value), and thus we may expect to be able to obtain better values through a dedicated tuning phase; $(ii)$ the complexity of the data preparation phase is greatly reduced, since the decision tree is capable of directly exploiting textual information, allowing one to avoid the ngram preprocessing step, and the related a-priori choices of the ngram sizes, frequencies, and feature selection strategies altogether; $(iii)$ sequential patterns may be more resistant to noise in the data than bigrams and trigrams, as they are capable of skipping irrelevant words thanks to the \emph{maxGap} parameter; $(iv)$ finally, and most importantly, the constructed trees are much more understandable to domain experts (which are not computer scientists) than regression models.
As an example, in Figure \ref{alberosalutifinali} we show a very simple model that recognizes the presence of the tag \emph{greeting\_final} (in this case, just combining single words).

\begin{table*}[t]
\centering
\caption{Tagging performance with J48S, over Kaldi (K) and Google (G) transcriptions}
\label{tabellatreetag}
\begin{tabular}{lcccccccc}
\hline
& \multicolumn{2}{c}{\textbf{Accuracy}} 
& \multicolumn{2}{c}{\textbf{Precision}} 
& \multicolumn{2}{c}{\textbf{Recall}} 
& \multicolumn{2}{c}{\textbf{TNR}} \\ 
\textbf{Keyword} & \textbf{K} & \textbf{G} & \textbf{K} & \textbf{G} & \textbf{K} & \textbf{G} & \textbf{K} & \textbf{G}  \\
\hline
age                   & 0.964 &  0.957    & 0.935 &  0.964     & 0.793 & 0.985   & 0.991  & 0.821    \\
call\_permission      & 0.923 &  0.958    & 0.671 &  0.776     & 0.681 & 0.678   & 0.955  & 0.983    \\
duration\_info        & 0.971 &  0.955    & 0.891 &  0.821     & 0.828 & 0.677   & 0.988  & 0.985    \\
family\_unit          & 0.971 &  0.970    & 0.877 &  0.929     & 0.864 & 0.855   & 0.985  & 0.989    \\
greeting\_initial     & 0.973 &  0.974    & 0.924 &  0.921     & 0.827 & 0.910   & 0.992  & 0.986   \\
greeting\_final       & 0.988 &  0.990    & 0.965 &  0.989     & 0.916 & 0.899   & 0.996  & 0.999   \\
person\_identity      & 0.944 &  0.940    & 0.763 &  0.830     & 0.752 & 0.758   & 0.970  & 0.972   \\
privacy               & 0.992 &  0.995    & 0.955 &  0.990     & 0.955 & 0.960   & 0.995  & 0.999    \\
profession            & 0.984 &  0.974    & 0.897 &  0.893     & 0.906 & 0.844   & 0.991  & 0.989    \\
question\_1           & 0.990 &  0.992    & 0.985 &  0.972     & 0.927 & 0.946   & 0.998  & 0.997    \\
question\_2           & 0.980 &  0.990    & 0.914 &  0.963     & 0.883 & 0.956   & 0.991  & 0.995    \\
question\_3           & 0.983 &  0.992    & 0.919 &  0.974     & 0.919 & 0.948   & 0.991  & 0.997   \\
\hline
\textbf{Average}      & 0.972 & 0.974     & 0.891 &  0.919     & 0.854 & 0.868   & 0.987 &  0.976 \\
\hline
\end{tabular}
\end{table*}

\iffalse
\begin{table}[t]
\centering
\caption{Tagging performance with J48S.}
\label{tabellatreetag}
\begin{tabular}{lllll}
\hline
\textbf{Keyword name} & \textbf{Accuracy} & \textbf{Precision} & \textbf{Recall} & \textbf{TNR} \\
\hline
age                   & 0.9638            & 0.9353             & 0.7927          & 0.9912       \\
call\_permission      & 0.9226            & 0.6713             & 0.6809          & 0.9551       \\
duration\_info        & 0.9705            & 0.8908             & 0.8281          & 0.9877       \\
family\_unit          & 0.9714            & 0.8769             & 0.8636          & 0.9848       \\
greeting\_initial     & 0.9731            & 0.9244             & 0.8271          & 0.9915       \\
greeting\_final       & 0.9882            & 0.9646             & 0.9160          & 0.9963       \\
person\_identity      & 0.9444            & 0.7630             & 0.7518          & 0.9696       \\
privacy               & 0.9916            & 0.9550             & 0.9550          & 0.9954       \\
profession            & 0.9840            & 0.8969             & 0.9063          & 0.9908       \\
question\_1           & 0.9899            & 0.9845             & 0.9270          & 0.9981       \\
question\_2           & 0.9798            & 0.9138             & 0.8833          & 0.9906       \\
question\_3           & 0.9832            & 0.9194             & 0.9194          & 0.9906      \\
\hline
\textbf{Average} & 0.9719 & 0.8913 & 0.8543 & 0.9868 \\
\hline
\end{tabular}
\end{table}
\fi

\section{On the operational impact of the speech analytics process}
\label{busimpact}

The introduction of a speech analytics process in a contact center may have a really wide and deep impact. Improving customer experience, reducing costs, identifying selling opportunities, reducing attrition and churn, evaluating agents, and rising service quality are some of the sought-after advantages. Among them, the improvement of service quality and the overall evaluation of the workforce are very important business goals, and constitute the main driving factors of the present study. 

Most of the operational processes that take place in a contact center are quite complex and, in a modern company, widely based on business intelligence. 
The experimentation illustrated in the present work shows that with a relatively small effort it is possible to develop a fully fledged in-house speech analytics solution, the adoption of which may imply a refactoring of operational processes in order to get the best results in terms of service quality and costs control. 

In detail, Gap S.r.l.u.\ makes use of a call quality assurance process with two dimensions: \emph{external}, tailored to the needs of customers and final users, and \emph{internal}, pointing to operational issues like, for instance, agent management. The process consists of a series of steps:
\begin{enumerate}
\item \emph{call scope evaluation}, that is, determining if the call is pertinent to the service;
\item \emph{checking of mandatory contents}, e.g., checking  whether or not all the relevant questions of an outbound survey have been asked;
\item \emph{proper data collection and actions to be performed}, e.g., sending a payment reminder as one of the results of a call;
\item \emph{correct call outcome classification}, by which the call outcome manually set by the agent is validated;
\item \emph{evaluation of soft skills}, e.g., checking whether the agent has been polite during the call.
\end{enumerate}

Speech analytics clearly impacts on all the above steps, as the exploitation of conversation transcripts may potentially play a major role in each of the tasks. Moreover, all phases are susceptible of being automatized, and, by formulating suitable business rules, many automatic actions may be performed. Examples include signaling to the operational staff the most problematic calls and issuing specific training tasks to cope with the deficiencies emerged in the handling of the calls.

Highlighting some detail, in order to better identify relevant parts of the agent side conversation, Gap S.r.l.u.\ operational staff has organized the semantic tags discovered by the speech analytics process in a set of logical rules. The Boolean values of the rules identify in a simple way what is considered to be compliant and what is not, according to the service operational guidelines. As a further step, a partitioning of the tags into three groups helps the operational staff in identifying the severity level of the flaws discovered in a call.

Let us now give an example that shows how tags, logical rules, and severity levels work together in the process. The operational guidelines state that the initial greeting and the call permission items have to be evaluated together, and
the corresponding check is considered to be passed only if both of them are present. This can be represented by means of a simple logical rule: \emph{greeting\_initial AND call\_permission}. Moreover, the tag partitioning leads to establish the severity level of potentially missing tags. According to the guidelines, in our example the tags \emph{greeting\_initial} and \emph{call\_permission} are both categorized in the same set associated with the severity level 2 (mid-level, which means that the call has some problems, but the survey is still valid). Table \ref{tab:esempiruleslevel} presents some further concrete examples taken from the service operational guidelines.

\begin{table*}[t]
\centering
\caption{Correspondence among tags, logical rules, and severity levels}
\label{tab:esempiruleslevel}
\begin{tabular}{lll}
\hline
\textbf{Tag}                                                                                        & \textbf{Logical rule}                                                                                & \textbf{Severity}   \\
\hline
age, person\_identity
& age $\wedge$ person\_identity
& 1                     \\
greeting\_initial, \mbox{call\_permission}
& greeting\_initial $\wedge$ \mbox{call\_permission}
& 2 \\
question\_1, question\_2, \mbox{question\_3}
& question\_1 $\wedge$ question\_2 $\wedge$ \mbox{question\_3}
& 3                    \\
\hline
\end{tabular}
\end{table*}

To summarize, by combining semantic tags, logical rules, and partitioning, each single call may be analyzed in detail and a clear assessment pattern emerges. Such a method gives the operational staff the ability to gain solid insights in a faster way and with more precise results with respect to a manual call monitoring procedure. The supervising staff confirms an improvement of 15\% in the overall quality of the process with respect to the old procedure, which required a manual sampling of the calls to be monitored, followed by a completely handmade evaluation of their contents. Considering the time spent for the analysis, the gain is about 17\%. If we restrict out attention to the detection of surveys to be canceled for serious flaws committed by the agents, then the increment is sensible, with results from the field indicating a gain of more than 300\%.

It is worth to underline once more that, as shown, these results can be accomplished by relying on the Kaldi-based proprietary transcription model, confirming that the operational fallout is pretty good, despite its higher word error rate with respect to the Google Speech API solution. Moreover, the tested solution has virtually no costs except for the internal one and this is clearly a relevant benefit for a company. Observe also that no dedicated or advanced hardware is needed for its implementation.

As already pointed out, the Gap S.r.l.u.\  IT and operational staffs are evaluating further automatic steps that can be possibly added to the call quality assurance process. As an example, on the basis of the logical rules outcomes and the severity level of the flaws, a set of e-learning modules could be suggested to the involved agents. This would lead to an instance of what is known as the \lq\lq learning analytics cycle\rq\rq{} \citep{clow2012learning}, i.e., an ongoing (and almost fully automated) circular process involving learners (the agents), data (the transcribed speech), metrics (the speech analysis), and interventions (the e-learning and supervisoring). This seems to be a very effective and interesting evolution for the operational impact of the entire project.

\section{Conclusions}
\label{sec:conclusions}

In this work, the experimentation of a speech analytics process for an Italian contact center has been illustrated. It consists of two distinct phases: a transcription step, where agent side conversations are transformed to texts, and a tagging step, where semantic content is attached to the transcripts. 

As for the transcription phase, the feasibility of building an in-house solution, relying on a small amount of training data, has been confirmed, by effectively developing a model using the Kaldi framework and comparing its performance with those obtained from the application of Google Cloud Speech API. 

As for the tagging phase, the tested machine learning based strategies have proven themselves to be reliable, and more efficient than classic solutions based on regular expressions. 
Specifically, the decision tree inducing algorithm J48S, applied to the problem of tagging, has exhibited a performance comparable to classical, ngram based approaches, while relying on just a single, highly interpretable model and reducing the complexity of the data preprocessing phase.

Overall, based on the experimental evidence that we provided, we can safely claim that the considered speech analytics process can be built at a low-cost for the company, relying on in-house developed models that exploit already available hardware, with respect to both the transcription and the tagging phases.

\section*{Data and code availability}

\paragraph{Availability of data and material}
None published regarding the proprietary corpora.  The 
CLIPS corpus is available online at  \url{http://www.clips.unina.it/it/corpus.jsp}, while the QALL-ME corpus is available at \url{http://qallme.fbk.eu/}.

\paragraph{Code availability}
A more recent development of J48S, called J48SS, is available at \url{https://github.com/dslab-uniud/J48SS}. The Kaldi framework is available at \url{https://kaldi-asr.org/}.

\bibliographystyle{spbasic}   
\bibliography{./bibliography.bib}

\end{document}